\documentclass[11pt]{article}

\usepackage[final]{acl}

\usepackage{times}
\usepackage{latexsym}
\usepackage[T1]{fontenc}
\usepackage[utf8]{inputenc}
\usepackage{microtype}
\usepackage{inconsolata}
\usepackage{graphicx}
\graphicspath{{figures/}}

\usepackage{booktabs}
\usepackage{multirow}
\usepackage{array}
\usepackage{colortbl}
\usepackage{xcolor}
\usepackage{adjustbox}
\usepackage{amsmath}
\usepackage{amssymb}
\usepackage{enumitem}

\usepackage{tikz}
\usetikzlibrary{positioning,arrows.meta,shapes.geometric,fit,backgrounds,calc}
\usepackage[most]{tcolorbox}

\definecolor{aclnavy}{HTML}{1F3B57}
\definecolor{aclsteel}{HTML}{2C6F9B}
\definecolor{aclteal}{HTML}{2E8B8B}
\definecolor{aclorange}{HTML}{E07B39}
\definecolor{aclpurple}{HTML}{6B5B95}
\definecolor{acllight}{HTML}{EAF0F5}
\definecolor{aclrule}{HTML}{9AA7B8}
\definecolor{rowgray}{HTML}{F2F5F8}

\usepackage{listings}
\tcbuselibrary{listings,breakable,skins}
\definecolor{promptheader}{HTML}{3B6BA5}
\definecolor{promptbody}{HTML}{FBFCFE}
\definecolor{promptframe}{HTML}{C3CBD6}
\newtcblisting{promptbox}[1][]{%
  enhanced jigsaw, breakable, listing only,
  colback=promptbody, colframe=promptframe, boxrule=0.6pt, arc=1.5pt,
  left=5pt, right=5pt, top=3pt, bottom=3pt,
  colbacktitle=promptheader, coltitle=white, titlerule=0pt,
  fonttitle=\bfseries\footnotesize, toptitle=2.5pt, bottomtitle=2.5pt, lefttitle=6pt,
  listing options={basicstyle=\scriptsize\ttfamily, breaklines=true,
    breakatwhitespace=false, columns=fullflexible, keepspaces=true,
    showstringspaces=false, breakindent=0pt, aboveskip=2pt, belowskip=2pt},
  #1}

\newcolumntype{R}{>{\raggedleft\arraybackslash}p{0.9cm}}

\newcommand{\appref}[1]{Appendix~\ref{#1}}

\title{\includegraphics[height=1.1cm]{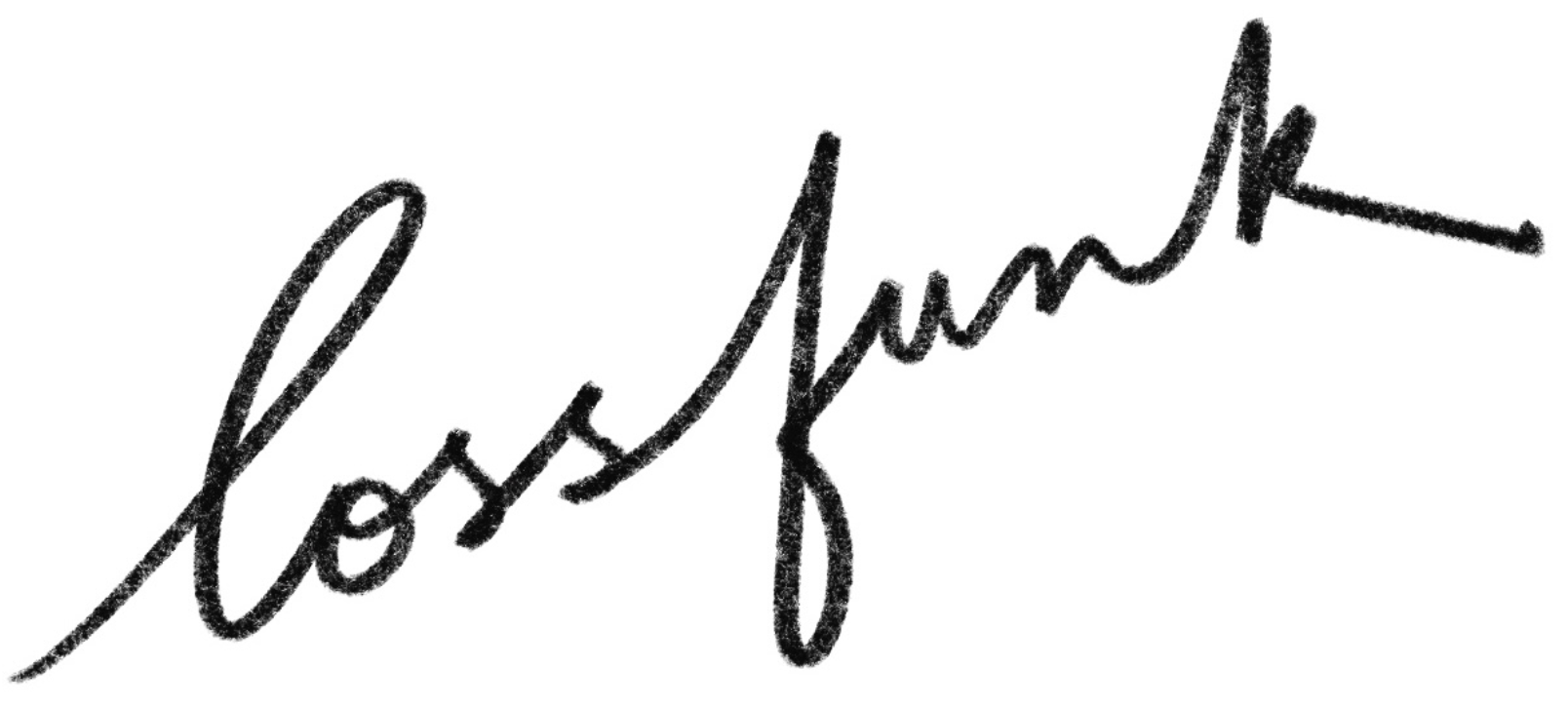}\\[6pt]
Do Vision--Language Models See or Guess? Measuring and Reducing\\ Textual-Prior Reliance with a Phrasing-Controlled Benchmark}

\author{
  Pratham Singla\textsuperscript{1,2} \quad Shivank Garg\textsuperscript{1,2} \quad Vihan Singh\textsuperscript{3} \quad Paras Chopra\textsuperscript{1} \\[4pt]
  \textsuperscript{1}Lossfunk \quad
  \textsuperscript{2}Indian Institute of Technology Roorkee \quad
  \textsuperscript{3}Raeth AI \\[2pt]
  \texttt{pratham\_s@me.iitr.ac.in} \quad \texttt{shivank\_g@mfs.iitr.ac.in} \\
  \texttt{vihan@raeth.ai} \quad \texttt{paras@lossfunk.com}
}

\begin{document}
\maketitle

\begin{abstract}
Vision-language models (VLMs) are increasingly deployed where answers must follow from what is in the image, yet they often answer from textual priors, the question's phrasing together with memorized world knowledge, rather than from the image itself, which inflates benchmark scores and yields confident but ungrounded answers. Existing benchmarks rarely isolate this behavior, since each image is usually paired with a single fixed question. To measure the reliance, we build a 540-image benchmark across six reasoning categories and generate four question variants over the same images, so that phrasing rather than image content is the controlled variable. The hardest variant is written directly from the image to minimize text leakage. We benchmark eleven VLMs spanning small open-weight models to large closed-source systems: every model degrades on the hardest variant, and open models fall furthest. Our central diagnostic is a no-image ablation, which collapses the open-weight models to their text-only floor (1 to 9 percent). Three further analyses, LLM-rated difficulty, low base-to-final textual similarity, and human re-annotation, corroborate genuine image-dependence. In-context exemplars that match how a variant was built recover the most accuracy, and GRPO post-training of a small VLM yields consistent gains across all four variants that transfer to a held-out out-of-distribution set. Textual-prior reliance is measurable and partly trainable away. We will release our dataset and code upon acceptance.
\end{abstract}
\begin{figure*}[t]
  \centering
  \includegraphics[width=0.95\linewidth]{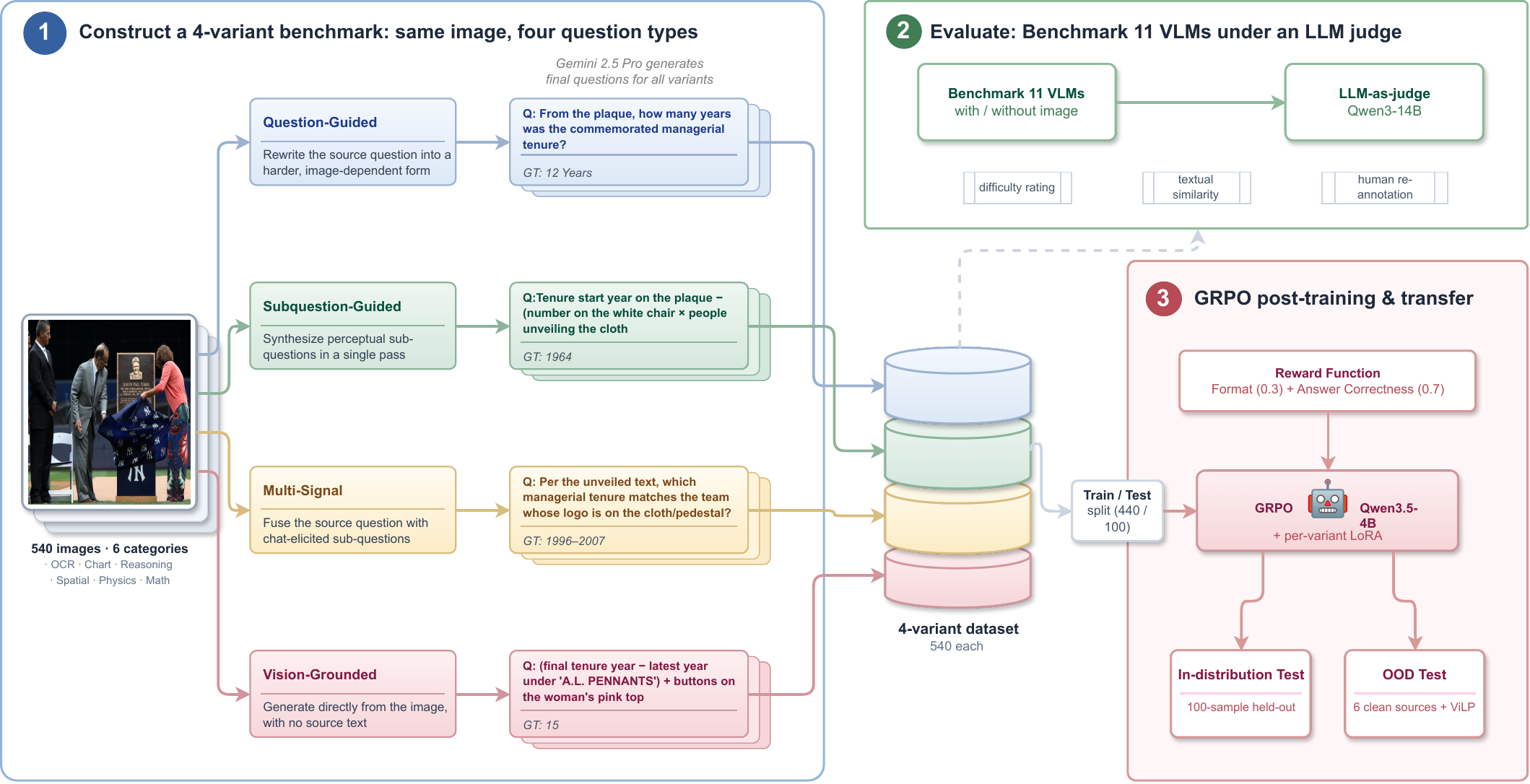}
  \caption{Overview of the pipeline. From each of 540 images across six reasoning categories,
  Gemini-2.5-Pro generates four question variants of the \emph{same} image (one colour each),
  differing only in the signal that conditions generation: the base question and answer
  (Question-Guided), single-pass perceptual sub-questions (Subquestion-Guided), chat-elicited
  sub-questions (Multi-Signal), or the image alone (Vision-Grounded). The worked example traces
  one image through all four. We use the resulting four-part benchmark to \textbf{(i)} evaluate
  eleven VLMs with and without the image (LLM judge: Qwen3-14B) and \textbf{(ii)} GRPO post-train
  Qwen3.5-4B with per-variant LoRA adapters, tested in- and out-of-distribution.}
  \label{fig:pipeline}
\end{figure*}

\section{Introduction}
\label{sec:intro}
Vision-Language Models (VLMs) now sit behind a widening range of multimodal applications, from
document understanding to visual question answering, and their reported accuracy on standard benchmarks suggests that they read images competently. A growing body of work complicates this picture. VLMs fail on simple perceptual tasks that humans solve trivially
\citep{rahmanzadehgervi2024blind,tong2024eyeswideshut}, hallucinate objects and relations that
are absent from the image \citep{guan2024hallusionbench}, and frequently produce the same answer when the image is withheld entirely \citep{vilp2025}. These behaviors point to a common cause: models lean on \emph{textual priors}, the surface form of the question together with knowledge memorized during pretraining, instead of grounding their answers in pixels. In deployed settings this is a silent failure mode, since an ungrounded answer looks identical to a correct one until the image actually matters. The phenomenon echoes long-standing findings in visual question answering, where language priors and distribution shift between training and test questions let models guess correctly without looking \citep{goyal2017vqav2,agrawal2018vqacp}.
The difficulty is that standard accuracy does not separate the two ways a model can be right.
A high score conflates ``the model saw the image'' with ``the model already knew the answer from the question text,'' because most benchmark questions are answerable, or nearly so, from phrasing and world knowledge alone. Harder benchmarks raise the ceiling on reasoning difficulty \citep{lu2024mathvista,yue2024mmmu} but were not designed to isolate image-dependence: a question can be both hard and answerable from text, so a low score there is equally consistent with weak reasoning and with weak perception. What is missing is a controlled way to vary how much a question reveals while holding the underlying image fixed, so that any change in accuracy can be attributed to the question rather than to the visual content.
We close this gap with a benchmark of 540 images spanning six reasoning categories, over which we generate four question variants over the same images (Figure~\ref{fig:pipeline}). Because all variants share the same images, phrasing is the controlled lever: an easier variant rewords a source question, while the hardest variant, \emph{Vision-Grounded}, is written directly from the image and so minimizes the text leakage that lets a model answer without looking. We diagnose reliance with a no-image ablation that re-runs each open model on the questions alone, and we then ask whether the reliance can be reduced through GRPO post-training rather than treated as a fixed property of a model. Across eleven VLMs, every model degrades on the image-derived Vision-Grounded variant, and open models fall hardest (10--16\% versus 27--38\% for proprietary systems). With the image withheld, open-model accuracy collapses to its text-only floor, and GRPO post-training then reduces
this reliance both in- and out-of-distribution.

We make the following contributions:
\begin{itemize}[leftmargin=1.2em,itemsep=2.5pt,topsep=2pt,parsep=0pt]
  \item \textbf{A phrasing-controlled, image-fixed benchmark.} 540 images across six reasoning
  categories, each paired with four question variants generated over the \emph{same} image, so that question phrasing rather than visual content is the controlled variable.
  \item \textbf{A broad eleven-model evaluation.} We quantify the open--proprietary accuracy gap and show that the hardest, image-derived variant is the hardest column for every model.
  \item \textbf{A no-image diagnostic with triangulated validation.} Withholding the image is our
  central test of reliance; LLM-rated difficulty, base-to-final textual similarity, and human
  re-annotation corroborate that the questions genuinely require vision.
  \item \textbf{Recovering and reducing the deficit.} In-context exemplars matching how a variant was constructed recover the most accuracy, showing the deficit is missing grounding rather than
  ill-posedness; GRPO post-training of a small VLM then reduces textual-prior reliance across all four variants, with gains that transfer to a provenance-clean out-of-distribution set.
\end{itemize}

\section{Related Work}
\label{sec:related}
\paragraph{Language priors and blind baselines.}
The tendency of VQA models to exploit textual shortcuts rather than image content was identified
early: \citet{goyal2017vqav2} showed that models trained on VQA v1 answered correctly even from
question text alone, motivating a balanced split that penalised yes/no guessing;
\citet{agrawal2018vqacp} further demonstrated that distribution shifts in question-answer
correlations expose severe over-reliance on language priors.
More recent work reveals the same pathology in modern VLMs.
\citet{rahmanzadehgervi2024blind} documented cases where large vision-language models fail
elementary tasks solvable by any sighted person, while \citet{tong2024eyeswideshut} showed that
CLIP-based models share systematic blind spots that propagate into VLMs built on top of them.
\citet{guan2024hallusionbench} demonstrated that VLMs frequently hallucinate answers inconsistent
with the provided image, and \citet{vilp2025} constructed a textual-prior probe whose items are
deliberately answerable from text alone to measure the degree of vision bypass.
A no-image (text-only) baseline unifies these findings:
withholding the image isolates exactly how much a model relies on the question's phrasing.
Unlike ViLP's hand-built text-answerable items, we vary phrasing across four variants over a fixed
540-image set and pair this with a no-image ablation at scale, treating image-dependence as a
primary experimental variable rather than a secondary diagnostic.

\paragraph{VLM evaluation benchmarks.}
Many benchmarks evaluate VLM capability across diverse tasks: MMMU and MMMU-Pro
stress college-level disciplinary reasoning \citep{yue2024mmmu,yue2024mmmupro};
MM-Vet, SEED-Bench, and MMBench probe instruction-following and compositional perception
\citep{yu2023mmvet,li2023seedbench,liu2024mmbench};
MMStar filters out items solvable without any image \citep{chen2024mmstar};
MathVista and MathVision target mathematical visual reasoning \citep{lu2024mathvista,wang2024mathvision};
ChartQA tests structured chart interpretation \citep{masry2022chartqa}.
These benchmarks measure a wide range of competencies, but each image typically appears with a
single fixed question, so it is difficult to separate model difficulty attributable to visual
content from difficulty attributable to question phrasing.
By pairing every image with four variants at differing linguistic distances from the base question,
our benchmark directly controls for phrasing and measures how accuracy changes as the question
demands more genuine visual processing.

\paragraph{Automatic and synthetic question generation.}
Automatic generation of hard evaluation items has grown common as models approach
human performance on hand-curated benchmarks.
\citet{li2024autobencher} showed that strong language models can author adversarial questions
tailored to probe model weaknesses, and the LLM-as-judge paradigm \citep{zheng2023judging}
provides a scalable substitute for human graders when ground-truth matching is ambiguous.
Our pipeline uses Gemini-2.5-Pro to generate four question variants per image through four
distinct prompting strategies (base-question rewrite, sub-question decomposition, multi-signal
fusion, and direct image-only generation), then validates each variant three ways: Claude-rated difficulty
scores, Claude-rated base-to-final textual similarity, and human re-annotation via a purpose-built
web application.
This multi-layer validation distinguishes our approach from prior synthetic benchmarks that rely
on generation alone and treat LLM judgment as sufficient.

\paragraph{RL post-training for reasoning.}
Reinforcement learning from verifiable rewards has become a practical route to stronger multi-step
reasoning; \citet{shao2024deepseekmath} introduced GRPO, which replaces the value network with
group-relative reward baselines, and LoRA \citep{hu2022lora} makes such adaptation cheap.
We apply GRPO with LoRA to Qwen3.5-4B, one adapter per question variant, to test whether
textual-prior reliance can be reduced in-distribution and whether the gains transfer to
provenance-clean out-of-distribution images.

\section{A Four-Variant Image-Dependent Benchmark}
\label{sec:benchmark}
We curate 540 images from 21 publicly available benchmarks spanning six reasoning
categories: OCR (72), Chart/Graphic Understanding (120), Common Sense \& Physics (79),
Spatial \& Scene Reasoning (94), Visual Reasoning (155), and Visual Math Reasoning (20).
Source benchmarks include
MathVista~\citep{lu2024mathvista},
ChartQA~\citep{masry2022chartqa},
TextVQA~\citep{singh2019textvqa},
DocVQA~\citep{mathew2021docvqa},
GQA~\citep{hudson2019gqa},
OK-VQA~\citep{marino2019okvqa},
ScienceQA~\citep{lu2022scienceqa},
AI2D~\citep{kembhavi2016ai2d},
RealWorldQA~\citep{realworldqa2024}, and
CLEVR-Math~\citep{lindstrom2022clevrmath},
among others; the full source-to-category mapping is given in \appref{app:dataset}.
Images were sampled randomly from each source subject to category balance targets.
To rule out cross-source duplicates, every image is fingerprinted with a perceptual hash,
and any near-duplicate pair is removed before the final pool is assembled.

Each image is reused across four question variants, which are
summarised in Table~\ref{tab:variants} and whose category
breakdown appears in Table~\ref{tab:categories}.
The first variant, \textbf{Question-Guided}, takes the original source question
alongside the image and asks Gemini-2.5-Pro~\citep{comanici2025gemini25} to rewrite it into a harder, more
image-dependent form.
The second, \textbf{Subquestion-Guided}, generates its perceptual sub-questions in a single
pass and synthesizes them into the final question, so that the phrasing reflects lower-level
perceptual decompositions rather than the original wording.
The third, \textbf{Multi-Signal}, instead elicits the sub-questions turn-by-turn in a
multi-turn chat (each conditioned on prior turns) and fuses them with the source question
and a chat summary.
The fourth and hardest variant, \textbf{Vision-Grounded}, provides Gemini-2.5-Pro with
only the image and no textual context, demanding that the generation rely entirely on
visual content; the result is a question whose wording shares no lineage with any
existing annotation.
Verbatim generation prompts for all four strategies are reproduced in \appref{app:prompts}.

Fixing the image set while varying question phrasing is the central design choice.
Because every variant describes the same scene, differences in model accuracy across
variants cannot be attributed to changes in visual content; they must instead reflect
how much each phrasing allows a model to answer from memorised text patterns rather
than from the image.
This logic is validated by the no-image ablation described in \S\ref{sec:diagnosis},
where withholding the image from the six open-weight models collapses accuracy to
1--9\%, confirming that the generated questions are not answerable from text alone.

We use the original source questions as a reference tier (the \textbf{Base Question}),
giving a directly comparable easier condition over the same 540 images.
All four generated variants and the base reference are drawn from the same underlying
image pool, so Table~\ref{tab:mainacc} can place proprietary and open models on a
common axis without confounds from image-domain shift.

Finally, the 540 images are partitioned into a 440-image training set and a 100-image
held-out test split (stratified by category, seed 42), which feeds the GRPO post-training
study in \S\ref{sec:grpo}.

\begin{table}[t]
\centering
\small
\setlength{\tabcolsep}{6pt}
\begin{tabular}{@{}lp{4.7cm}@{}}
\toprule
\textbf{Variant} & \textbf{Generation strategy} \\
\midrule
Question-Guided    & Rewrite the source question into a harder, image-dependent form \\
Subquestion-Guided & Synthesize perceptual sub-questions in a single pass \\
Multi-Signal       & Fuse the source question with chat-elicited sub-questions \\
Vision-Grounded    & Generate directly from the image, with no source text \\
\bottomrule
\end{tabular}
\caption{The four question variants generated over the same 540 images, ordered by decreasing
reliance on the source text. A 100-sample test split is held out for the GRPO study
(\S\ref{sec:grpo}).}
\label{tab:variants}
\end{table}

\begin{table}[t]
\centering
\small
\setlength{\tabcolsep}{6pt}
\begin{tabular}{@{}lr@{}}
\toprule
\textbf{Category} & \textbf{Samples} \\
\midrule
OCR & 72 \\
Chart / Graphic Understanding & 120 \\
Common Sense \& Physics & 79 \\
Spatial \& Scene Reasoning & 94 \\
Visual Reasoning & 155 \\
Visual Math Reasoning & 20 \\
\midrule
\textbf{Total (per variant)} & \textbf{540} \\
\bottomrule
\end{tabular}
\caption{The six reasoning categories spanned by the benchmark.}
\label{tab:categories}
\end{table}

\section{Evaluation Setup}
\label{sec:setup}
We evaluate eleven VLMs divided into two groups.
The \textbf{proprietary/API} group comprises Claude Sonnet 4.6~\citep{anthropic2026claudesonnet46},
Gemini 2.5 Pro~\citep{comanici2025gemini25} and Gemini 3.1 Flash-Lite~\citep{gdm2026gemini31flashlite},
and GPT-5 mini~\citep{openai2025gpt5}.
The \textbf{open-weight} group consists of
Qwen3.5-397B-A17B and Qwen3.5-9B~\citep{qwenteam2026qwen35},
InternVL3.5-8B~\citep{wang2025internvl35},
LLaVA-OneVision-1.5 8B and LLaVA-OneVision-1.5 4B~\citep{an2025llavaonevision15},
Llama-3.2-11B-Vision~\citep{grattafiori2024llama3},
and Phi-4 Multimodal~\citep{abouelenin2025phi4mm}.

All models receive each question together with its image (\textit{with-image} condition).
For the six smaller open-weight models (Qwen3.5-9B, InternVL3.5-8B, both LLaVA-OV
sizes, Llama-3.2-11B-Vision, Phi-4 Multimodal), we additionally run a
\textit{no-image} ablation in which the image is withheld and only the question text
is presented; this isolates the contribution of visual information to each model's
accuracy.

\paragraph{Evaluation metric.}
We report accuracy as the fraction of items judged correct by an LLM-as-judge.
Following \citet{zheng2023judging}, we use Qwen3-14B~\citep{yang2025qwen3} as an automated judge that issues
a binary yes/no equivalence verdict for each (ground-truth answer, model response) pair.
Answers extracted from models that emit chain-of-thought output enclosed in
\texttt{<think>} tags undergo lenient extraction: if the closing \texttt{</think>} tag
is absent, the full completion is used as the candidate answer rather than discarding
the response.
This protocol is applied uniformly across all models and both the with-image and
no-image conditions.
GRPO post-training and OOD evaluation, which use the same judge and the same extraction
rule, are described separately in \S\ref{sec:grpo}.

\section{Do VLMs See or Guess?}
\label{sec:diagnosis}
\paragraph{With-image accuracy.}
Table~\ref{tab:mainacc} reports per-variant accuracy with the image present, and
Appendix~\ref{app:percat} visualizes the same numbers as a model-by-variant heatmap. Two patterns
hold across all eleven VLMs. First, every model is substantially less accurate on the four
generated variants than on the base question: Qwen3.5-9B, for example, drops from 69.3\% on
the base question to 42.0\% on Vision-Grounded, and Phi-4 Multimodal from 49.8\% to 10.0\%.
Second, Vision-Grounded, the variant written directly from the image, is uniformly the hardest
column. The open--proprietary gap that is modest on the base question widens sharply here: open
models score 10--16\% on Vision-Grounded (Phi-4 10.0\%, LLaVA-OV-1.5 4B 12.6\%, InternVL3.5-8B
15.9\%) while proprietary models hold 27--38\% (Gemini 3.1 Flash-Lite 27.3\%, Qwen3.5-397B-A17B
38.4\%). The ordering is monotone in source text: variants built from more of the original text
(Multi-Signal) stay closer to base accuracy, while the variant with no source text falls
furthest. This base-to-variant drop is the first sign that the generated questions resist
textual shortcuts, since the underlying images are unchanged and only the phrasing varies. The
drop alone is only suggestive, however; accuracy could fall simply because the new questions are
harder for reasons unrelated to vision. We therefore turn to an ablation that isolates the
image's contribution directly.

\begin{figure}[t]
  \centering
  \includegraphics[width=\columnwidth]{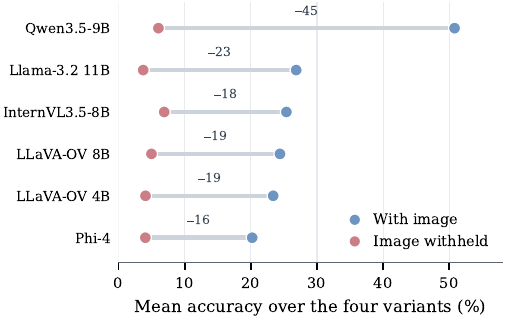}
  \caption{Removing the image collapses open-model accuracy to its text-only floor (mean over the
  four variants). The gap estimates the image's contribution; it ranges from 16 to 45 points.}
  \label{fig:noimage}
\end{figure}

\paragraph{The no-image ablation.}
The decisive test is to ask the same questions with the image withheld: a model that was reading
the image should now fail, whereas one exploiting textual priors should retain much of its
accuracy. Running this control on the six open-weight models collapses their accuracy to roughly
1--9\% (Figure~\ref{fig:noimage}). The floor holds for every question type, including the easy
base reference: the same models answer the base question 40--69\% of the time with the image but
only 5--9\% without it. This leaves no substantial text-only-solvable subset of the kind that
inflates scores on unbalanced datasets~\citep{goyal2017vqav2,chen2024mmstar}, and image-dependence
is a property of the whole benchmark rather than of its hardest variants alone. Because the question text is held fixed and
only the image is removed, this image-contribution gap cannot be attributed to phrasing or
memorized world knowledge, and the accuracy that survives without the image is negligible; this
establishes directly what the base-to-variant drop only suggested: these questions cannot be
answered from textual priors. Per-model contribution gaps, the
benchmark-balance comparison, and the reason the control runs on the open models only are reported
in \appref{app:fulltables}. The remaining analyses corroborate image-dependence three independent
ways.

\begin{table*}[t]
\centering
\small
\setlength{\tabcolsep}{6pt}
\begin{tabular}{@{}lccccc@{}}
\toprule
\textbf{Model} & \textbf{Base Question} & \textbf{\shortstack[c]{Question-\\Guided}} & \textbf{\shortstack[c]{Subquestion-\\Guided}} & \textbf{\shortstack[c]{Multi-\\Signal}} & \textbf{\shortstack[c]{Vision-\\Grounded}} \\
\midrule
Claude Sonnet 4.6        & 64.6 & 41.7 & 47.8 & 54.3 & 31.6 \\
Gemini 2.5 Pro           & 74.3 & 46.1 & 53.3 & 58.5 & 35.9 \\
Gemini 3.1 Flash-Lite    & 58.9 & 37.0 & 41.8 & 43.7 & 27.3 \\
GPT-5 mini               & 61.1 & 46.3 & 49.4 & 49.4 & 33.3 \\
\midrule
Qwen3.5-397B-A17B        & 64.8 & 48.4 & 54.6 & 54.8 & 38.4 \\
Qwen3.5-9B               & 69.3 & 45.7 & 56.8 & 58.7 & 42.0 \\
InternVL3.5-8B           & 50.2 & 20.6 & 28.9 & 36.1 & 15.9 \\
LLaVA-OneVision-1.5 8B   & 47.2 & 18.7 & 29.8 & 35.9 & 13.2 \\
LLaVA-OneVision-1.5 4B   & 45.7 & 16.5 & 29.8 & 34.6 & 12.6 \\
Llama-3.2 11B Vision     & 39.8 & 23.3 & 31.3 & 37.6 & 15.2 \\
Phi-4 Multimodal         & 49.8 & 19.1 & 22.6 & 29.1 & 10.0 \\
\bottomrule
\end{tabular}
\caption{LLM-judged accuracy (\%) with image, on the base question and the four generated
variants. Top block: proprietary/API models; bottom block: open-weight models.
Vision-Grounded is uniformly the hardest variant; open models fall to 10--16\% on it while
proprietary models stay at 27--38\%.}
\label{tab:mainacc}
\end{table*}

\section{Are the Questions Really Image-Dependent?}
\label{sec:validation}
\begin{figure}[t]
  \centering
  \includegraphics[width=\columnwidth]{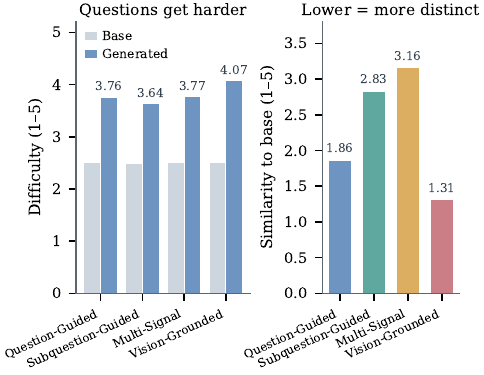}
  \caption{The generated questions are markedly harder than the base questions they extend
  (left: mean difficulty, base$\rightarrow$final, Claude 1--5) and far less textually derivable
  from them (right: similarity to the base question, where lower means more distinct).
  Vision-Grounded is both the hardest and the most distinct.}
  \label{fig:diffsim}
\end{figure}

\paragraph{Difficulty.}
We first ask whether the generated variants are objectively harder than the base questions they
extend. Using Claude as a rater on a 1--5 scale, the base questions average roughly 2.50, while
all four variants rate higher: 3.76 for Question-Guided, 3.64 for Subquestion-Guided, 3.77 for
Multi-Signal, and 4.07 for Vision-Grounded (Figure~\ref{fig:diffsim}). The ordering tracks the
with-image results of \S\ref{sec:diagnosis}: Vision-Grounded, the variant no model handled well
and the one written without any source text, also receives the largest jump in rated difficulty
(+1.57 over base). The variants are thus harder by an independent measure, not merely harder for
the models.

\paragraph{Base-to-final textual similarity.}
Higher difficulty would be uninteresting if the hard questions were simply reworded base
questions, since a model could then answer them by the same textual route. To rule this out, we
measure how textually similar each generated question is to its base question, again with Claude
on a 1--5 scale where lower means more distinct and therefore less answerable from the base
text. The variants that reuse more source text are more similar (Question-Guided 1.86,
Subquestion-Guided 2.83, Multi-Signal 3.16), while Vision-Grounded is the most distinct at 1.31
(Figure~\ref{fig:diffsim}). Low similarity means the hardest questions are not paraphrases of
the easy base question, so their difficulty cannot be a phrasing artifact carried over from the
source; combined with the difficulty scores, this places Vision-Grounded as both the hardest
and the least text-derivable variant, consistent with its behavior under the no-image ablation.

\paragraph{Human re-annotation.}
Finally, we re-annotate the generated answers with human reviewers to confirm that the questions
are answerable and that the ground truth is sound. Each item was independently judged by a panel
of three annotators (two to three judgments per item); inter-annotator agreement is strong on the
text-derived variants (Krippendorff's $\alpha=0.85$--$0.91$; \appref{app:human}). The fraction of
generated answers a reviewer
corrected rises steeply with variant difficulty: 21.9\% for Question-Guided (118/540), 11.7\%
for Subquestion-Guided (63), and 11.9\% for Multi-Signal (64), but 42.8\% for Vision-Grounded
(231) (Table~\ref{tab:human}). The hardest variant draws by far the most correction, in line
with its higher difficulty and lower similarity, and the human-verified answers tighten the
ground truth used to score every model. These three analyses share a caveat: the difficulty and
similarity scores come from a single LLM judge (Claude) rather than independent human ratings, so
they should be read as corroborating the no-image ablation, not as standalone ground truth.
Difficulty in particular is rated as \emph{predicted model failure}, so it is not fully
independent of the accuracies it tracks. The
human re-annotation, by contrast, rests on human judgments and so anchors the others. Together
they give three converging signals that the generated questions are harder than, and genuinely
distinct from, the base questions, reinforcing the ablation's conclusion that answering them
requires the image.

\begin{table}[t]
\centering
\small
\setlength{\tabcolsep}{5pt}
\begin{tabular}{@{}lccc@{}}
\toprule
\textbf{Variant} & \textbf{Samples} & \textbf{Changed} & \textbf{\% changed} \\
\midrule
Question-Guided   & 540 & 118 & 21.9 \\
Subquestion-Guided & 540 &  63 & 11.7 \\
Multi-Signal         & 540 &  64 & 11.9 \\
Vision-Grounded   & 540 & 231 & \textbf{42.8} \\
\bottomrule
\end{tabular}
\caption{Human re-annotation: fraction of generated answers corrected by reviewers. The
hardest variant draws the most intervention, consistent with difficulty rather than
generation noise.}
\label{tab:human}
\end{table}

\section{Recovering Accuracy with the Right Grounding}
\label{sec:icl}
The no-image ablation shows that our questions cannot be answered from text alone, but it
leaves open a second possibility: that the questions are simply unanswerable, ill-posed rather
than merely difficult. If instead the gap reflects missing grounding or decomposition, then
supplying a model with the right kind of intermediate evidence at inference time should recover
much of the lost accuracy. We test this with in-context exemplars whose form matches how each
variant was constructed. The construction recipe for a variant is, in effect, a hypothesis about
what scaffolding the question needs; giving the model that same scaffolding as a worked example
should help most when the example type matches the variant.

We run this on the three question-derived variants and exclude Vision-Grounded, which is written
directly from the image and has no source question from which to draw an exemplar. We pair each
variant with three exemplar types, each presented with its ground-truth answer: the original base
question, the single-pass sub-questions for Subquestion-Guided, and the chat-style
sub-questions for Multi-Signal. Every model sees the same image and target question and differs only in which exemplar
type accompanies it, so any change in accuracy is attributable to the exemplar rather than to the
question or the image.

\begin{figure}[t]
  \centering
  \includegraphics[width=\columnwidth]{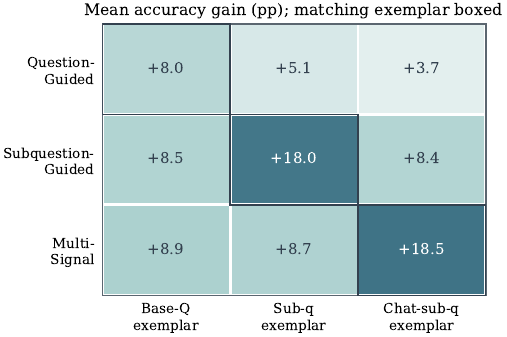}
  \caption{In-context exemplars help most when their \emph{type matches} how a variant was
  constructed (boxed diagonal): base-question exemplars for Question-Guided, sub-question
  exemplars for Subquestion-Guided, and chat sub-question exemplars for Multi-Signal. Cells
  show the mean accuracy gain (pp) over the no-exemplar baseline, averaged across the eleven
  models; Vision-Grounded is excluded as it has no source exemplar.}
  \label{fig:icl}
\end{figure}

The matching exemplar yields the largest mean gain for each variant
(Figure~\ref{fig:icl}). Supplying the base question and its answer raises Question-Guided
accuracy by $8.0$ points; supplying the single-pass sub-questions raises Subquestion-Guided by $18.0$ points;
and supplying the chat-style sub-questions raises Multi-Signal by $18.5$ points. The alignment
between the exemplar that helps most and the recipe that built the variant holds across the three
cases, and the full per-model tables are reported in Appendix~\ref{app:fulltables}.

This result clarifies what the earlier diagnostics measure. The questions are answerable once the
model is given the visual grounding and the intermediate reasoning their construction assumes, so
these variants expose a failure of grounding and decomposition, not unanswerability. Models that score at their text-only floor without the image, and well below
base-question accuracy with it, recover a large share of that accuracy when handed an exemplar of
the right type. Textual-prior reliance therefore reflects a failure to assemble the visual
evidence a question demands, which motivates the targeted training we turn to next.

\section{Mitigating Reliance with GRPO}
\label{sec:grpo}
The in-context result shows that the right grounding recovers accuracy at inference time. We now
ask whether the same behavior can be trained into the model's weights, so that it grounds answers
in the image without being handed an exemplar.

\begin{figure}[t]
  \centering
  \includegraphics[width=\columnwidth]{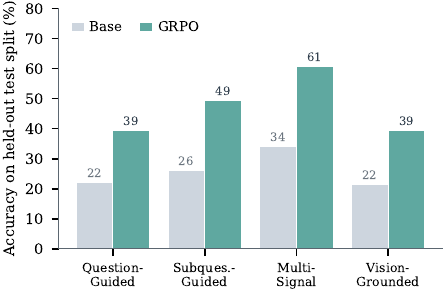}
  \caption{GRPO post-training of Qwen3.5-4B improves all four variants on the held-out
  in-distribution test split (per-variant LoRA adapters vs.\ the untrained base).}
  \label{fig:grpo}
\end{figure}

\paragraph{In-distribution.}
We fine-tune Qwen3.5-4B with GRPO \citep{shao2024deepseekmath}, learning one LoRA adapter
\citep{hu2022lora} per variant on that variant's training split and rewarding correct, properly
formatted answers; the vision tower is frozen and only language layers are updated, with full
hyperparameters in \appref{app:grpo}. Each adapter is evaluated on the 100-sample held-out test
split of its own variant. All four adapters improve over the untrained base
(Figure~\ref{fig:grpo}; per-variant values in Table~\ref{tab:grpo}): Question-Guided rises from $22.2$ to $39.3$
($+17.1$ points), Subquestion-Guided from $26.1$ to $49.4$ ($+23.3$), Multi-Signal from $34.0$ to
$60.7$ ($+26.7$), and Vision-Grounded from $21.5$ to $39.3$ ($+17.8$). We read these numbers as
a consistent gap-closing pattern rather than as precise effect sizes. The test split is small,
several reasoning categories contribute only a handful of items each, and accuracy is computed
with a lenient \texttt{</think>} extraction that falls back to the full completion when the model
omits the closing tag, which tends to favor recall over precision. Part of the in-distribution
gain may therefore reflect the answer formatting the reward directly optimizes; the
out-of-distribution transfer below, which the reward does not target, is the de-confounded
evidence. The pattern that every variant
moves in the same direction under matched per-variant training is the load-bearing claim; the
exact magnitudes should be read with these caveats and are revisited in \S\ref{sec:limitations}.

\begin{figure}[t]
  \centering
  \includegraphics[width=\columnwidth]{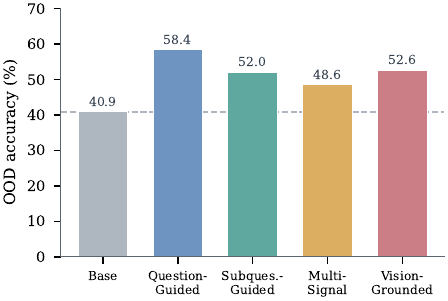}
  \caption{Out-of-distribution transfer on the 200-sample OOD set (six provenance-clean
  sources plus a ViLP probe). Every adapter improves over the base model; the Question-Guided
  adapter generalizes best.}
  \label{fig:ood}
\end{figure}

\paragraph{Out-of-distribution generalization.}
Gains on a model's own training distribution can reflect format adaptation as much as improved
grounding, so we test transfer on questions drawn from sources the model never saw in training. We
assemble a 200-question open-ended OOD set from six provenance-clean benchmarks, none of which
overlaps the 21 sources used to build our benchmark, together with a 40-item probe from ViLP
\citep{vilp2025} that targets textual-prior reliance directly. The base model and all four
adapters are evaluated on the identical questions, so differences isolate the effect of training.
The base model reaches $40.9\%$, and every adapter improves on it; the Question-Guided adapter
transfers best at $58.4\%$ ($+17.5$ points), with the remaining adapters between $48.6\%$ and
$52.6\%$ (Figure~\ref{fig:ood}). The adapter trained on the simplest image-grounded variant
therefore generalizes furthest to unseen sources, consistent with the in-context finding that
base-question grounding is broadly useful.

Taken together, the in-distribution and OOD results indicate that textual-prior reliance is
partially trainable: a small VLM rewarded for grounded, correct answers improves on every
variant it is trained for and transfers part of that gain to questions from new sources.

\section{Conclusion}
\label{sec:discussion}
The no-image ablation is the load-bearing diagnostic in this work: with the image withheld, open-model accuracy collapses to its text-only floor, so the generated questions cannot be answered from phrasing or memorized world knowledge alone. Three further analyses point the same way. Rated difficulty rises across the variants, textual similarity to the base question falls, and human corrections climb sharply on the hardest one. The in-context result sharpens the interpretation: once a model is given exemplars matching how a variant was built, the questions become answerable, so the deficit is missing grounding rather than ill-posedness. GRPO post-training then trains part of the reliance away. Rewarding a small VLM for grounded, correctly formatted answers raises accuracy on all four variants in-distribution, and the gains carry over to an out-of-distribution set the reward never targeted, without unfreezing the vision encoder or scaling the backbone.

Two directions stand out: a tolerance-aware scorer for numeric and spatial answers would credit responses that exact match rejects, and scaling beyond a single four-billion-parameter backbone would test whether the effect grows with capacity. Textual-prior reliance is real, it is measurable with a controlled image ablation, and it is partially trainable away.

\section*{Limitations}
\label{sec:limitations}
Our post-training results are promising in direction but modest in magnitude, and several
constraints bound their reach. GRPO is tested on a single 4-billion-parameter backbone
(Qwen3.5-4B) with LoRA adapters and a frozen vision encoder, evaluated on a 100-sample held-out
split whose per-category counts are too small for reliable per-category claims, so we report
aggregate variant trends only. The GRPO scores also use a lenient answer extraction that falls
back to the full completion when the model omits the trained \texttt{</think>} tag, which raises
base accuracy and compresses the reported gains relative to strict scoring; the released
completions keep the strict figures reconstructible. Finally, the difficulty and
textual-similarity ratings come from a single LLM judge (Claude) and serve only as corroborating
evidence for image-dependence, not as ground truth, read alongside the judge-independent no-image
ablation.

\section*{Ethics Statement}
All 540 images come from existing, publicly released vision-language benchmarks
(Appendix~\ref{app:dataset}); we do not redistribute the images themselves but release the
generated question variants, ground-truth answers, and human-corrected annotations together with
identifiers that let users retrieve each image from its original source. Generated answers were
reviewed and, where needed, corrected by human annotators through a structured interface, and the
benchmark is intended for evaluation and model-improvement research on VLM textual-prior reliance,
not for consequential decisions about individual users or systems. Because the reported numbers
reflect a single evaluation round with one LLM judge (Qwen3-14B) and a fixed prompt, they
characterize behaviour on this benchmark rather than definitive capability rankings.

\bibliography{custom}

\appendix
\section{Generation, Judging, and Scoring Prompts}
\label{app:prompts}

This appendix reproduces the prompts used throughout the pipeline. The four question variants
are produced by a common template that differs only in what context the generator is given.
Vision-Grounded receives the image alone; we reproduce its prompt in full, then list the lines
that distinguish the text-conditioned variants. All question-variant generation used Gemini-2.5-Pro (intermediate sub-questions used Gemini-2.0-Flash); difficulty and
similarity used Claude Sonnet~4.6; answer judging used Qwen3-14B.

\begin{promptbox}[title=Vision-Grounded question generation (Gemini-2.5-Pro; image only)]
You are an expert AI assistant designed to craft **extremely challenging evaluation questions
for Vision-Language Models (VLMs)**. Your task is to analyze an image and generate a
**difficult, single query-based question** that requires genuine multimodal reasoning.

OBJECTIVE:
Create a **high-quality, image-dependent difficult question** along with its **complete ground
truth answer**. The question must challenge a model's ability to perform **complex visual
reasoning** -- not solvable by text-only language models.

GUIDELINES FOR QUESTION GENERATION:
1. Visual-Only Answerability: The question must require direct, careful inspection of the image.
   Do NOT describe or hint at visual elements in the question explicitly. Make the question
   answerable ONLY by examining the image carefully.
2. Complex Reasoning Required: go beyond surface-level understanding -- visual trends, spatial
   relations, visual logic, graphical interpretation, mathematical computation from visual
   features, comparative analysis, or multi-step deduction. Avoid hallucinations.
3. Precise and Focused Query: a concise, single query, not a multi-part exam. Combine multiple
   reasoning needs into one challenging prompt with a clear, unambiguous answer.
4. High Difficulty Level: hard enough to differentiate strong and weak VLMs; not solvable through
   pattern matching, guessing, or world knowledge alone.
5. Objective & Verifiable Framing: a clear, correct answer; avoid subjective formulations; focus
   on facts, counts, relationships, measurements, or logical conclusions verifiable from the image.

GUIDELINES FOR GROUND TRUTH ANSWER:
- Provide detailed step-by-step reasoning grounded in visual elements, then end with:
  `Final Answer: your_answer_here`
- Keep the answer verifiable from the image, unambiguous, complete, with units if applicable.

RESPONSE FORMAT (strictly follow):
Final Question: [the challenging query -- do not describe the image explicitly]
Ground Truth Answer:
[detailed step-by-step reasoning based on visual analysis]
Final Answer: [final answer]
\end{promptbox}

\noindent The three text-conditioned variants reuse the template above with one substituted
instruction. \textbf{Question-Guided} replaces guideline~1 with ``Build upon the provided
original question: transform or enhance it into a more challenging, image-dependent query.''
\textbf{Subquestion-Guided} and \textbf{Multi-Signal} replace it with ``Synthesize the provided
sub-questions (randomly sample five if more are given) into one compact, unified query,'' with
Multi-Signal additionally receiving the base question and a chat-style summary.
The two variants differ in how their sub-questions are obtained: Subquestion-Guided produces all
five sub-questions in a single pass, whereas Multi-Signal elicits them one at a time in a multi-turn
chat, each conditioned on the preceding turns.

\begin{promptbox}[title=Question difficulty rating (Claude Sonnet 4.6; scale 1--5)]
Please rate the difficulty of the following visual question for the model to answer correctly,
given the associated image. The difficulty reflects how likely the model is to make mistakes,
misinterpret the image or question, or fail to produce a complete and correct response.

There are five levels of question difficulty:
1 - Very Easy: visually and linguistically simple; almost certainly answered correctly.
2 - Easy: generally easy; minor visual or linguistic misunderstandings possible.
3 - Moderate: may partially struggle with visual reasoning or question understanding.
4 - Hard: requires complex visual reasoning, fine-grained perception, or multi-step inference.
5 - Very Hard: highly challenging visually or conceptually; very likely answered incorrectly.

Only output the score (a number), do not give any explanation.
Use the correct answer as the reference for evaluation.

[question begin]
{question}
[question end]
Correct Answer: {correct_answer}
\end{promptbox}

\begin{promptbox}[title=Base--final question similarity rating (Claude Sonnet 4.6; scale 1--5)]
You are an expert at analyzing the semantic similarity between questions in the context of
visual question answering tasks. Rate the similarity between two questions that are both related
to the provided image, reflecting how closely they align in intent, meaning, and the information
they seek to extract from the image.

SIMILARITY RATING SCALE (1-5):
1 - Not Similar: entirely different aspects of the image, minimal or no overlap.
2 - Slightly Similar: some thematic overlap but diverge in core intent or focus.
3 - Moderately Similar: noticeable shared intent, but differ in scope or specificity.
4 - Very Similar: closely aligned, essentially the same information; minor phrasing differences.
5 - Nearly Identical: semantically equivalent.

- Output ONLY a single number from 1 to 5; do NOT provide explanation.

Question 1: {question1}
Question 2: {question2}
\end{promptbox}

\begin{promptbox}[title=LLM-as-judge answer equivalence (Qwen3-14B)]
You are an expert evaluator comparing answers for accuracy.

EVALUATION CRITERIA:
1. For factual answers (numbers, dates, names): must match exactly or be semantically equivalent.
2. For descriptive answers: check semantic similarity, key concepts, and factual accuracy.
3. For yes/no questions: both answers must have the same conclusion.
4. Respond with ONLY "Yes" or "No" based on whether the groundtruth and predicted answers are
   the same or equivalent.

INPUT:
Groundtruth: {ground_truth}
Predicted: {predicted}

OUTPUT:
\end{promptbox}

\section{GRPO Training Details}
\label{app:grpo}

We post-train Qwen3.5-4B with one LoRA adapter per variant using Unsloth~\citep{daniel2023unsloth}
and the TRL \texttt{GRPOTrainer}~\citep{vonwerra2022trl}, optimizing the GRPO
objective~\citep{shao2024deepseekmath}. Each adapter is trained on its own variant's 440-image
training split and evaluated on the corresponding 100-sample held-out test split.
Table~\ref{tab:hyperparams} lists the configuration. We use
sequence-level importance sampling (GSPO)~\citep{zheng2025gspo} with the Dr.GRPO
loss~\citep{liu2025drgrpo} and no KL penalty to the reference model.
Training was performed on a single NVIDIA RTX~A6000 GPU (48\,GB); each per-variant adapter required
approximately 80 GPU-hours ($\approx$320 GPU-hours total across the four adapters).

\begin{table}[h]
\centering
\small
\setlength{\tabcolsep}{6pt}
\begin{tabular}{@{}ll@{}}
\toprule
\textbf{Component} & \textbf{Setting} \\
\midrule
Base model & Qwen3.5-4B (4-bit) \\
Adapters & one LoRA per variant \\
LoRA rank / $\alpha$ / dropout & 16 / 16 / 0.1 \\
Trainable modules & language layers (vision frozen) \\
Learning rate & $5\times10^{-6}$ \\
Optimizer / schedule & adamw\_8bit / cosine \\
Warmup ratio & 0.05 \\
Generations per prompt & 8 \\
Sampling temperature & 1.0 \\
Importance sampling & sequence-level (GSPO) \\
Loss / KL $\beta$ & Dr.GRPO / 0 \\
Epochs & 3 \\
Max sequence length & 8192 \\
Precision & bf16 \\
Hardware & NVIDIA RTX A6000 (48\,GB) \\
\bottomrule
\end{tabular}
\caption{GRPO hyperparameters.}
\label{tab:hyperparams}
\end{table}

\begin{figure}[h]
  \centering
  \includegraphics[width=\columnwidth]{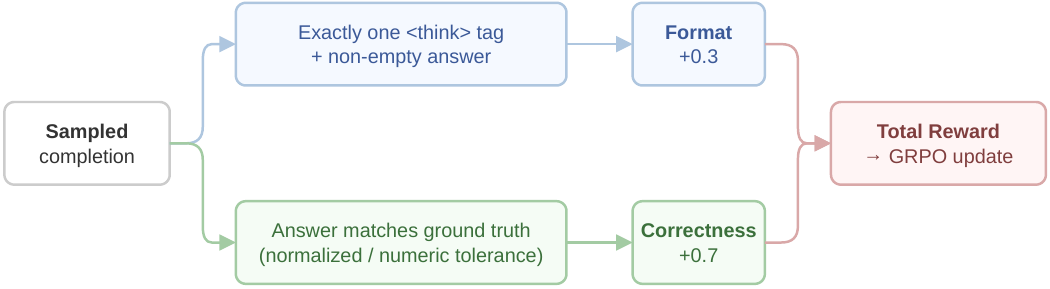}
  \caption{The GRPO reward. Each sampled completion earns a \emph{format} term (a well-formed
  reasoning tag plus a non-empty answer) and a \emph{correctness} term (the answer matching the
  ground truth under normalized or numeric-tolerant comparison); their sum drives the policy update.}
  \label{fig:reward}
\end{figure}

\paragraph{Reward.} The total reward is the sum of two binary terms applied to each sampled
completion. A \emph{format} reward of $0.3$ is granted when the completion contains exactly one
\texttt{</think>} tag followed by a non-empty answer. A \emph{correctness} reward of $0.7$ is
granted when the extracted final answer matches the ground truth under normalized exact match,
a relative numeric tolerance ($<10^{-3}$), or a trailing-number match for numeric targets;
normalization lowercases, strips punctuation and articles, and maps spelled-out numbers to digits.
Figure~\ref{fig:reward} summarizes the two reward terms.

\begin{table}[h]
\centering
\small
\setlength{\tabcolsep}{6pt}
\begin{tabular}{@{}lccc@{}}
\toprule
\textbf{Variant} & \textbf{Base} & \textbf{GRPO} & \textbf{$\Delta$ (pp)} \\
\midrule
Question-Guided   & 22.2 & 39.3 & +17.1 \\
Subquestion-Guided & 26.1 & 49.4 & +23.3 \\
Multi-Signal         & 34.0 & 60.7 & \textbf{+26.7} \\
Vision-Grounded   & 21.5 & 39.3 & +17.8 \\
\bottomrule
\end{tabular}
\caption{GRPO vs.\ base on the in-distribution 100-sample test split (lenient
\texttt{</think>} extraction). The split is small and per-category counts are tiny; we report
the gap-closing \emph{pattern} rather than single-variant magnitudes (\S\ref{sec:limitations}).}
\label{tab:grpo}
\end{table}

\begin{figure*}[t]
  \centering
  \includegraphics[width=\textwidth]{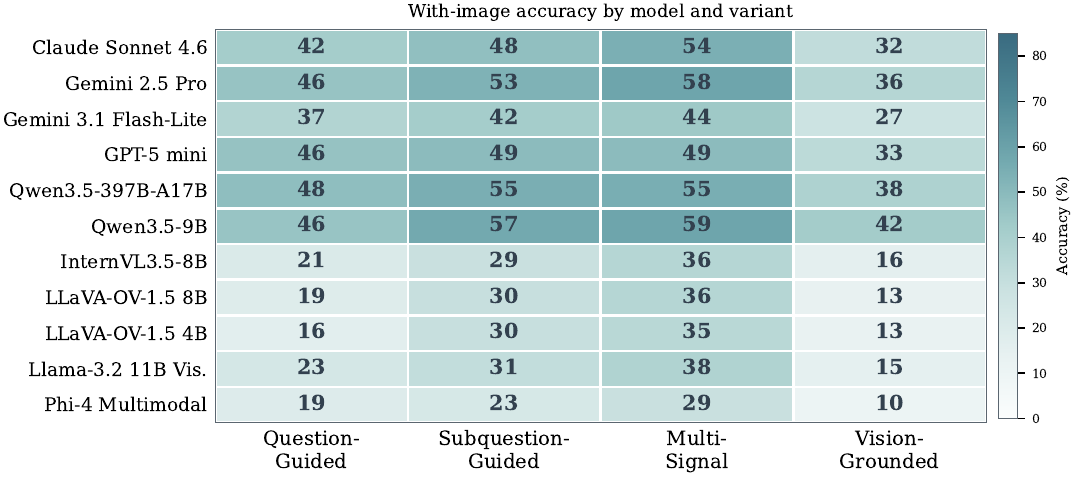}
  \caption{With-image accuracy by model and variant (overall). The Vision-Grounded column is
  consistently lowest, and the open-weight block (lower rows) trails the proprietary block.}
  \label{fig:heatmap}
\end{figure*}

\section{Extended Related Work}
\label{app:related}

Our in-distribution benchmark and out-of-distribution set together span the breadth of existing
VLM suites, drawn from disjoint sources and de-duplicated by perceptual hash so that no image
recurs across them: OCR and document understanding
\citep{singh2019textvqa,mathew2021docvqa,mathew2022infographicvqa,mishra2019ocrvqa}; charts,
tables, and diagrams \citep{masry2022chartqa,wang2024charxiv,kembhavi2016ai2d,kim2024tablevqa};
counting and compositional reasoning \citep{paiss2023countbench,hudson2019gqa,lindstrom2022clevrmath};
knowledge and science \citep{marino2019okvqa,lu2022scienceqa}; spatial and real-world perception
\citep{liao2024qspatial,realworldqa2024}; commonsense-defying images \citep{bittonguetta2023whoops};
and visual mathematics \citep{lu2024mathvista,wang2024mathvision,zhang2024mathverse}. Relative to
holistic suites \citep{yue2024mmmu,yue2024mmmupro,liu2024mmbench,li2023seedbench,chen2024mmstar},
our contribution is not a new task distribution but a controlled manipulation of question phrasing
over a fixed image set, paired with a no-image ablation that isolates the visual contribution.

A parallel line of work expands VLM evaluation along axes orthogonal to ours. Some suites push raw
difficulty until current models score near zero \citep{roberts2025zerobench}, while others isolate a
single competence in depth, such as low-level visual perception \citep{fu2024blink}, physical-world
understanding \citep{chow2025physbench}, visual logical reasoning \citep{xiao2024logicvista}, or text
localization and reading \citep{fu2025ocrbenchv2}; holistic frameworks instead aggregate many datasets
to report breadth, robustness, and safety in one place \citep{lee2024vhelm}. Closest to our concern
are benchmarks that probe whether a model truly uses the image, such as diagnostic hallucination and
visual-illusion tests \citep{guan2024hallusionbench}, visual-prior probes \citep{vilp2025}, and
blind-baseline analyses that recover answers without pixels
\citep{goyal2017vqav2,tong2024eyeswideshut,rahmanzadehgervi2024blind}. These efforts broaden coverage
and surface failure modes, but each varies the task or the image while leaving phrasing uncontrolled;
our four-variant design holds the image fixed and varies only how the question is posed, so that the
accompanying no-image ablation attributes the resulting gaps to image use rather than to question
difficulty.

\section{Extended Dataset Details}
\label{app:dataset}

The 540 images are drawn from 21 publicly available benchmarks and partitioned into the six
categories of Table~\ref{tab:categories}. Sources contributing to each category include OCR
(TextVQA, DocVQA, InfographicVQA); Chart/Graphic (ChartQA, AI2D, TableVQA-Bench, CountBenchQA);
Common Sense \& Physics (OpenSpaces, ConflictVQA, ScienceQA, a real/fake image set); Spatial \&
Scene (SpaceLLaVA/vqasynth, COCO validation, RealWorldQA); Visual Reasoning (MM-Vet, SEED-Bench,
JourneyDB, the ``VLMs are blind'' probes, and additional reasoning items); and Visual Math
(MathVista with its IQTest, PaperQA, and FunctionQA subsets, MathVision, and CLEVR-Math).
Each image is fingerprinted with a perceptual hash and near-duplicate pairs are removed so that no
image recurs across sources or between the benchmark and the OOD set.

\begin{table}[h]
\centering
\small
\setlength{\tabcolsep}{5pt}
\begin{tabular}{@{}llr@{}}
\toprule
\textbf{Category} & \textbf{OOD source} & \textbf{$N$} \\
\midrule
OCR              & OCR-VQA          & 27 \\
Chart/Graphic    & CharXiv          & 27 \\
Common Sense     & WHOOPS!          & 27 \\
Spatial \& Scene & Q-Spatial-Bench  & 27 \\
Visual Reasoning & VisualPuzzles    & 26 \\
Visual Math      & MathVerse        & 26 \\
Textual prior    & ViLP probe       & 40 \\
\midrule
\textbf{Total}   &                  & \textbf{200} \\
\bottomrule
\end{tabular}
\caption{Out-of-distribution evaluation set: OCR-VQA~\citep{mishra2019ocrvqa},
CharXiv~\citep{wang2024charxiv}, WHOOPS!~\citep{bittonguetta2023whoops},
Q-Spatial-Bench~\citep{liao2024qspatial}, VisualPuzzles~\citep{visualpuzzles2025},
MathVerse~\citep{zhang2024mathverse}, and a ViLP probe~\citep{vilp2025}. All six benchmark
sources are provenance-clean: none of their images overlaps the 21 sources used to build the
in-distribution benchmark.}
\label{tab:oodsources}
\end{table}

\section{Human Annotation Protocol and Inter-Annotator Agreement}
\label{app:human}

Generated answers were reviewed through a custom web application (a FastAPI backend with a React
frontend backed by a relational store). Each task presented an annotator with the image, the
generated question, and the model-proposed answer; the annotator marked the answer correct or
incorrect, recorded a confidence rating, and supplied a corrected answer when needed. Each item was
assigned to a panel of three annotators and received two to three independent judgments in practice.
Final labels were resolved by semantic-similarity consensus voting: candidate answers were grouped
by meaning, the largest group's share defined a confidence score, and ties were broken in favor of
high-confidence corrections. We report the resulting correction rates in Table~\ref{tab:human}.

\paragraph{Inter-annotator agreement.}
We assess reliability on the binary correctness judgment (whether the model-proposed answer is
correct) that every annotator recorded. Because the number of judgments per item varies, we report
Krippendorff's $\alpha$ (nominal)~\citep{krippendorff2011alpha}, which is defined for any number of
raters and tolerates missing judgments, alongside the observed pairwise agreement; both are computed
over all available judgments (Table~\ref{tab:iaa}). Agreement on the three text-derived variants is
strong ($\alpha = 0.85$--$0.91$, observed agreement above $98\%$, and mean pairwise Cohen's $\kappa$
of $0.84$--$0.96$), and pooled across all variants $\alpha = 0.86$ at $98.3\%$ observed agreement.

\begin{table}[h]
\centering
\small
\setlength{\tabcolsep}{8pt}
\begin{tabular}{@{}lrcc@{}}
\toprule
\textbf{Variant} & \textbf{$N$} & \textbf{\% agree} & \textbf{$\alpha$} \\
\midrule
Question-Guided    & 540 & 98.1 & 0.91 \\
Subquestion-Guided & 540 & 98.3 & 0.86 \\
Multi-Signal       & 540 & 98.5 & 0.85 \\
Vision-Grounded    & 540 & 98.3 & 0.86 \\
\midrule
\textbf{Overall}   & \textbf{2160} & \textbf{98.3} & \textbf{0.86} \\
\bottomrule
\end{tabular}
\caption{Inter-annotator agreement on the binary correctness judgment, per variant and pooled.
Items received two to three independent judgments; $N$ is the number of items with at least two
judgments, \% agree is the mean pairwise observed agreement, and $\alpha$ is Krippendorff's nominal
coefficient.}
\label{tab:iaa}
\end{table}

\section{Per-Category Results (With Image)}
\label{app:percat}

Figure~\ref{fig:heatmap} visualizes overall with-image accuracy, and
Tables~\ref{tab:percat-base}--\ref{tab:percat-direct} give the per-category breakdown for each
generated variant. Per-category counts are small (Visual Math $N=20$ in particular), so individual
cells are noisy; the consistent pattern is that Vision-Grounded is hardest and the open--proprietary
gap is widest there.

\begin{table*}[t]
\centering
\small
\setlength{\tabcolsep}{3pt}
\begin{tabular}{@{}lcccccc@{}}
\toprule
\textbf{Model} & \textbf{OCR} & \textbf{Chart/Graphic} & \textbf{Common Sense} & \textbf{Spatial} & \textbf{Visual Reasoning} & \textbf{Visual Math} \\
\midrule
Claude Sonnet 4.6      & 66.7 & 71.7 & 53.2 & 31.9 & 38.7 & 65.0 \\
Gemini 3.1 Flash-Lite  & 62.5 & 58.3 & 50.6 & 33.0 & 38.7 & 40.0 \\
GPT-5 mini             & 66.7 & 51.7 & 50.6 & 30.9 & 38.7 & 55.0 \\
Qwen3.5-397B-A17B      & 69.4 & 70.0 & 60.8 & 37.2 & 52.3 & 75.0 \\
Qwen3.5-9B             & 62.5 & 77.5 & 54.4 & 37.2 & 45.8 & 70.0 \\
InternVL3.5-8B         & 15.3 & 11.7 & 31.6 & 26.6 & 21.3 & 15.0 \\
LLaVA-OV-1.5 8B & 9.7  & 10.0 & 31.6 & 17.0 & 22.6 & 30.0 \\
LLaVA-OV-1.5 4B & 22.2 & 10.0 & 25.3 & 13.8 & 16.8 & 10.0 \\
Llama-3.2 11B Vision   & 25.0 & 31.7 & 24.1 & 16.0 & 20.6 & 20.0 \\
Phi-4 Multimodal       & 13.9 & 5.0  & 27.9 & 18.1 & 25.2 & 45.0 \\
\bottomrule
\end{tabular}
\caption{Question-Guided: per-category accuracy (\%) with image.}
\label{tab:percat-base}
\end{table*}

\begin{table*}[t]
\centering
\small
\setlength{\tabcolsep}{3pt}
\begin{tabular}{@{}lcccccc@{}}
\toprule
\textbf{Model} & \textbf{OCR} & \textbf{Chart/Graphic} & \textbf{Common Sense} & \textbf{Spatial} & \textbf{Visual Reasoning} & \textbf{Visual Math} \\
\midrule
Claude Sonnet 4.6      & 73.6 & 70.0 & 65.8 & 53.2 & 61.3 & 75.0 \\
Gemini 3.1 Flash-Lite  & 59.7 & 48.3 & 58.2 & 46.8 & 51.0 & 50.0 \\
GPT-5 mini             & 73.6 & 57.5 & 64.6 & 50.0 & 57.4 & 60.0 \\
Qwen3.5-397B-A17B      & 76.4 & 65.8 & 69.6 & 52.1 & 60.0 & 90.0 \\
Qwen3.5-9B             & 72.2 & 79.2 & 65.8 & 51.1 & 62.6 & 85.0 \\
InternVL3.5-8B         & 30.6 & 5.8  & 43.0 & 33.0 & 38.7 & 10.0 \\
LLaVA-OV-1.5 8B & 29.2 & 7.5  & 51.9 & 28.7 & 38.7 & 15.0 \\
LLaVA-OV-1.5 4B & 33.3 & 6.7  & 49.4 & 24.5 & 40.6 & 20.0 \\
Llama-3.2 11B Vision   & 40.3 & 30.0 & 45.6 & 24.5 & 27.7 & 10.0 \\
Phi-4 Multimodal       & 19.4 & 5.8  & 29.1 & 28.7 & 31.0 & 15.0 \\
\bottomrule
\end{tabular}
\caption{Subquestion-Guided: per-category accuracy (\%) with image.}
\label{tab:percat-chat}
\end{table*}

\begin{table*}[t]
\centering
\small
\setlength{\tabcolsep}{3pt}
\begin{tabular}{@{}lcccccc@{}}
\toprule
\textbf{Model} & \textbf{OCR} & \textbf{Chart/Graphic} & \textbf{Common Sense} & \textbf{Spatial} & \textbf{Visual Reasoning} & \textbf{Visual Math} \\
\midrule
Claude Sonnet 4.6      & 80.6 & 76.7 & 63.3 & 46.8 & 56.8 & 75.0 \\
Gemini 3.1 Flash-Lite  & 70.8 & 67.5 & 63.3 & 46.8 & 52.3 & 55.0 \\
GPT-5 mini             & 75.0 & 59.2 & 64.6 & 45.7 & 63.2 & 65.0 \\
Qwen3.5-397B-A17B      & 86.1 & 69.2 & 63.3 & 46.8 & 60.0 & 90.0 \\
Qwen3.5-9B             & 77.8 & 80.8 & 74.7 & 55.3 & 60.7 & 80.0 \\
InternVL3.5-8B         & 50.0 & 17.5 & 48.1 & 29.8 & 41.9 & 35.0 \\
LLaVA-OV-1.5 8B & 44.4 & 21.7 & 49.4 & 31.9 & 39.4 & 30.0 \\
LLaVA-OV-1.5 4B & 41.7 & 20.0 & 46.8 & 28.7 & 40.0 & 35.0 \\
Llama-3.2 11B Vision   & 47.2 & 40.0 & 44.3 & 23.4 & 38.7 & 20.0 \\
Phi-4 Multimodal       & 33.3 & 8.3  & 43.0 & 21.3 & 38.1 & 50.0 \\
\bottomrule
\end{tabular}
\caption{Multi-Signal: per-category accuracy (\%) with image.}
\label{tab:percat-mini}
\end{table*}

\begin{table*}[t]
\centering
\small
\setlength{\tabcolsep}{3pt}
\begin{tabular}{@{}lcccccc@{}}
\toprule
\textbf{Model} & \textbf{OCR} & \textbf{Chart/Graphic} & \textbf{Common Sense} & \textbf{Spatial} & \textbf{Visual Reasoning} & \textbf{Visual Math} \\
\midrule
Claude Sonnet 4.6      & 52.8 & 67.5 & 53.2 & 35.1 & 32.3 & 65.0 \\
Gemini 2.5 Pro      & 48.6 & 70.0 & 49.4 & 39.4 & 45.2 & 75.0 \\
Gemini 3.1 Flash-Lite  & 38.9 & 35.8 & 36.7 & 29.8 & 25.8 & 60.0 \\
GPT-5 mini             & 50.0 & 61.7 & 46.8 & 44.7 & 40.6 & 70.0 \\
Qwen3.5-397B-A17B      & 55.6 & 66.7 & 58.2 & 44.7 & 47.1 & 65.0 \\
Qwen3.5-9B             & 50.0 & 75.0 & 44.3 & 37.2 & 31.6 & 70.0 \\
InternVL3.5-8B         & 12.5 & 7.5  & 22.8 & 17.0 & 18.1 & 30.0 \\
LLaVA-OV-1.5 8B & 13.9 & 2.5  & 21.5 & 10.6 & 17.4 & 20.0 \\
LLaVA-OV-1.5 4B & 9.7  & 6.7  & 13.9 & 14.9 & 14.8 & 25.0 \\
Llama-3.2 11B Vision   & 11.1 & 17.5 & 24.1 & 11.7 & 11.6 & 25.0 \\
Phi-4 Multimodal       & 12.5 & 2.5  & 15.2 & 11.7 & 9.0  & 25.0 \\
\bottomrule
\end{tabular}
\caption{Vision-Grounded: per-category accuracy (\%) with image.}
\label{tab:percat-direct}
\end{table*}

\section{Full In-Context, No-Image, and OOD Tables}
\label{app:fulltables}

Table~\ref{tab:imagecontrib} reports image-contribution gaps (with the base reference shown for
comparison) and Table~\ref{tab:noimage} the full no-image accuracies.
Tables~\ref{tab:icl-base}--\ref{tab:icl-mini} give the per-model in-context results summarized in
Figure~\ref{fig:icl}.

\paragraph{Per-model detail.}
The image-contribution gap, the per-model mean gain from adding the image back, spans 16--45\,pp
and peaks for Qwen3.5-9B at 44.8\,pp; its single largest drop is on Subquestion-Guided, from
56.8\% to 6.1\% (a 50.7\,pp fall). Because this control runs only on the six open models, we read
the open--proprietary gap on Vision-Grounded (\S\ref{sec:diagnosis}) as a capability difference
rather than a measured difference in reliance. We did not run the no-image condition on the
proprietary API models: the effect is already clear from the open models, and API inference was
constrained by budget.

\paragraph{Benchmark balance.}
The ablation also serves as a balance check. Earlier visual-question-answering datasets contained
many questions answerable from language priors alone, which inflated reported accuracy and
motivated balanced re-splits~\citep{goyal2017vqav2} and prior-shifted test sets that expose the
reliance~\citep{agrawal2018vqacp}; more recent audits find that a substantial fraction of items in
popular multimodal benchmarks remain solvable without the image~\citep{chen2024mmstar}. Our
benchmark shows no sizable text-only-solvable subset: the floor in Table~\ref{tab:noimage} holds for
every question type, and the base reference question, the easiest condition, loses 34--61\,pp when
the image is removed (Table~\ref{tab:imagecontrib}), comparable to the generated variants. We do
not balance complementary image pairs as in VQA v2; instead, image-grounded generation produces
questions that the ablation then confirms are image-dependent across the board.

\begin{table}[h]
\centering
\small
\setlength{\tabcolsep}{3pt}
\begin{tabular}{@{}lcccccc@{}}
\toprule
\textbf{Model} & \textbf{Base} & \textbf{Q.G} & \textbf{S.G} & \textbf{M.S} & \textbf{V.G} & \textbf{Mean} \\
\midrule
Qwen3.5-9B          & 60.6 & 40.3 & 50.7 & 50.0 & 38.1 & 44.8 \\
InternVL3.5-8B      & 43.5 & 15.2 & 19.8 & 26.7 & 12.2 & 18.5 \\
LLaVA-OV-1.5 8B     & 40.2 & 15.6 & 23.7 & 27.6 & 10.7 & 19.4 \\
LLaVA-OV-1.5 4B     & 40.1 & 14.3 & 25.5 & 27.0 & 10.4 & 19.3 \\
Llama-3.2 11B Vis.  & 34.4 & 21.9 & 27.2 & 30.9 & 12.6 & 23.2 \\
Phi-4 Multimodal    & 44.6 & 15.9 & 18.1 & 23.3 &  7.0 & 16.1 \\
\bottomrule
\end{tabular}
\caption{Image contribution (with $-$ without image, pp). The Base column is the reference
question; it drops 34--61\,pp when the image is removed, comparable to the generated variants,
so even the easiest condition requires the image. Mean is over the four generated variants
(Q.G: Question-Guided, S.G: Subquestion-Guided, M.S: Multi-Signal, V.G: Vision-Grounded).}
\label{tab:imagecontrib}
\end{table}

\begin{table}[h]
\centering
\small
\setlength{\tabcolsep}{3.5pt}
\begin{tabular}{@{}lccccc@{}}
\toprule
\textbf{Model} & \textbf{Base} & \textbf{Q.G} & \textbf{S.G} & \textbf{M.S} & \textbf{V.G} \\
\midrule
Qwen3.5-9B          & 8.7 & 5.4 & 6.1 & 8.7 & 3.9 \\
InternVL3.5-8B      & 6.7 & 5.4 & 9.1 & 9.4 & 3.7 \\
LLaVA-OV-1.5 8B     & 7.0 & 3.1 & 6.1 & 8.3 & 2.4 \\
LLaVA-OV-1.5 4B     & 5.6 & 2.2 & 4.3 & 7.6 & 2.2 \\
Llama-3.2 11B Vis.  & 5.4 & 1.5 & 4.1 & 6.7 & 2.6 \\
Phi-4 Multimodal    & 5.2 & 3.1 & 4.4 & 5.7 & 3.0 \\
\bottomrule
\end{tabular}
\caption{No-image (text-only) accuracy (\%) for the six open models. Every column, including the
base reference question, sits at the same 1--9\% text-only floor, indicating no substantial
text-only-solvable subset.}
\label{tab:noimage}
\end{table}

\begin{table}[h]
\centering
\small
\setlength{\tabcolsep}{4pt}
\begin{tabular}{@{}lcccc@{}}
\toprule
\textbf{Model} & \textbf{Base} & \textbf{base-Q} & \textbf{sub-q} & \textbf{chat-sub-q} \\
\midrule
Claude Sonnet 4.6      & 41.7 & \textbf{53.5} & 50.6 & 45.8 \\
Gemini 2.5 Pro      & 46.1 & \textbf{60.6} & 55.8 & 53.0 \\
Gemini 3.1 Flash-Lite  & 37.0 & \textbf{44.3} & 41.1 & 40.6 \\
GPT-5 mini             & 46.3 & \textbf{57.8} & 52.6 & 53.0 \\
Qwen3.5-397B-A17B      & 48.4 & \textbf{56.3} & 50.2 & 52.2 \\
Qwen3.5-9B             & 45.7 & 57.0 & \textbf{59.2} & 55.1 \\
InternVL3.5-8B         & 20.6 & \textbf{34.1} & 32.5 & 28.4 \\
LLaVA-OV-1.5 8B        & 18.7 & \textbf{20.1} & 19.1 & 19.3 \\
LLaVA-OV-1.5 4B        & 16.5 & \textbf{21.1} & 19.6 & 18.0 \\
Llama-3.2 11B Vis.     & 23.3 & \textbf{25.0} & 23.7 & 24.4 \\
Phi-4 Multimodal       & 19.1 & \textbf{21.5} & 14.6 & 14.3 \\
\bottomrule
\end{tabular}
\caption{In-context learning on Question-Guided. Bold marks the best exemplar type per model;
the matching base-question exemplar wins for ten of eleven models.}
\label{tab:icl-base}
\end{table}

\begin{table}[h]
\centering
\small
\setlength{\tabcolsep}{4pt}
\begin{tabular}{@{}lcccc@{}}
\toprule
\textbf{Model} & \textbf{Base} & \textbf{base-Q} & \textbf{sub-q} & \textbf{chat-sub-q} \\
\midrule
Claude Sonnet 4.6      & 47.8 & 61.3 & \textbf{70.2} & 60.2 \\
Gemini 2.5 Pro      & 53.3 & 66.8 & \textbf{77.0} & 66.5 \\
Gemini 3.1 Flash-Lite  & 41.8 & 53.3 & \textbf{62.2} & 52.2 \\
GPT-5 mini             & 49.4 & 65.9 & \textbf{70.0} & 63.9 \\
Qwen3.5-397B-A17B      & 54.6 & 65.0 & \textbf{77.0} & 63.9 \\
Qwen3.5-9B             & 56.8 & 67.8 & \textbf{76.3} & 65.6 \\
InternVL3.5-8B         & 28.9 & 44.6 & \textbf{51.1} & 43.5 \\
LLaVA-OV-1.5 8B        & 29.8 & 29.3 & \textbf{41.4} & 30.4 \\
LLaVA-OV-1.5 4B        & 29.8 & 32.2 & \textbf{41.8} & 31.1 \\
Llama-3.2 11B Vis.     & 31.3 & 30.6 & \textbf{43.6} & 35.2 \\
Phi-4 Multimodal       & 22.6 & 23.2 & \textbf{33.0} & 26.1 \\
\bottomrule
\end{tabular}
\caption{In-context learning on Subquestion-Guided. The matching sub-question exemplar
wins for all eleven models.}
\label{tab:icl-chat}
\end{table}

\begin{table}[h]
\centering
\small
\setlength{\tabcolsep}{4pt}
\begin{tabular}{@{}lcccc@{}}
\toprule
\textbf{Model} & \textbf{Base} & \textbf{base-Q} & \textbf{sub-q} & \textbf{chat-sub-q} \\
\midrule
Claude Sonnet 4.6      & 54.3 & 63.3 & 64.3 & \textbf{74.8} \\
Gemini 2.5 Pro      & 58.5 & 69.3 & 68.5 & \textbf{78.0} \\
Gemini 3.1 Flash-Lite  & 43.7 & 60.7 & 62.8 & \textbf{70.7} \\
GPT-5 mini             & 49.4 & 68.5 & 68.3 & \textbf{73.9} \\
Qwen3.5-397B-A17B      & 54.8 & 67.6 & 66.6 & \textbf{79.6} \\
Qwen3.5-9B             & 58.7 & 67.2 & 68.7 & \textbf{76.3} \\
InternVL3.5-8B         & 36.1 & 48.0 & 47.0 & \textbf{56.5} \\
LLaVA-OV-1.5 8B        & 35.9 & 38.1 & 38.0 & \textbf{49.6} \\
LLaVA-OV-1.5 4B        & 34.6 & 38.3 & 35.1 & \textbf{45.9} \\
Llama-3.2 11B Vis.     & 37.6 & 41.9 & 40.6 & \textbf{50.6} \\
Phi-4 Multimodal       & 29.1 & 27.8 & 28.1 & \textbf{40.4} \\
\bottomrule
\end{tabular}
\caption{In-context learning on Multi-Signal. The matching chat-sub-question exemplar wins for all
eleven models.}
\label{tab:icl-mini}
\end{table}

\end{document}